\RequirePackage{fix-cm}
\documentclass[11pt]{uiuc_paper}
\pdfoutput=1
\usepackage{amsmath,amssymb,amsthm}
\usepackage{array,tabularx}
\usepackage{float,needspace}
\usepackage{flafter}
\usepackage{algorithm,algpseudocode}
\usepackage{xurl}
\graphicspath{{images/}}
\definecolor{surgeteal}{HTML}{167F80}
\definecolor{surgered}{HTML}{B94433}
\definecolor{surgegray}{HTML}{586777}
\definecolor{surgegreen}{HTML}{287A4D}
\definecolor{surgegrid}{HTML}{C9C6C0}
\newcommand{\surge}{\textsc{Surge}}
\newcommand{\FC}{\mathrm{FC}}
\newcommand{\Keep}{\mathcal{K}}
\newcommand{\Comp}{\mathcal{R}}

\newcolumntype{Y}{>{\raggedright\arraybackslash}X}

\hypersetup{
 pdftitle={Does Scaling Reinforcement Learning Really Require More Training?},
 pdfauthor={Bangji Yang, Jiajun Fan, Hongbo Ma, Ruihan Guo, Ge Liu},
 pdfsubject={Policy-space scaling from fixed reinforcement learning histories}
}

\title{Does Scaling Reinforcement Learning\\Really Require More Training?}
\author[1,*]{Bangji Yang}
\author[1,*]{Jiajun Fan}
\author[2,*]{Hongbo Ma}
\author[1,*]{Ruihan Guo}
\author[1,\textdagger]{Ge Liu}
\affiliation[1]{University of Illinois at Urbana-Champaign}
\affiliation[2]{Tsinghua University}
\contribution[*]{Equal contribution}
\contribution[\textdagger]{Corresponding author}
\abstract{Scaling reasoning typically spends more compute on reinforcement learning (RL) or on inference. We show that a completed RL training history can yield policies stronger than the checkpoints visited by its optimizer. We call this \emph{policy-space scaling}: expanding the deployable policy set accessible from a fixed RL history, without extending training or increasing per-query inference computation. We instantiate it with SURGE (Scaling Up RL Gradient-free via Eigenspace fusion). SURGE combines two checkpoints from the same RL run: a high-accuracy anchor and a competitive donor that generates shorter responses. It expresses both checkpoints as changes from their shared initialization, then spectrally decomposes the anchor's update to retain its dominant component and incorporate the donor's complementary component. With a fixed target for how much of the anchor update to retain, SURGE determines the block size from the weights without testing candidate policies. We evaluate two 1.5B mathematical-reasoning histories, DeepSeek and Nemotron, and one 7B coding history, OLMo. SURGE improves benchmark-average accuracy over both input checkpoints while using fewer reasoning tokens than the anchor. It reaches 54.17\% on DeepSeek AIME24 against a measured native maximum of 50.83\%, and 83.7\% on OLMo HumanEval+ against 82.8\%. These gains exceed the observed training curves. Geometric controls support the importance of RL-update structure beyond weight displacement or token reduction alone. Each constructed model runs as a single policy. Our findings identify stored RL history as a reusable scaling resource: the capability available from a training run need not end at its best checkpoint.
}
\date{\today}
\correspondence{\email{geliu@illinois.edu}}

\begin{document}
\maketitle
\section{Introduction}
Reinforcement learning with verifiable rewards has become a practical route to improving reasoning in language models \citep{shao2024deepseekmath,deepseek2025r1}. When performance is insufficient, the usual response is to continue optimization or allocate more computation to inference. Both extend a computation process. We ask whether the same training run already contains enough structure to construct a stronger policy. A training trajectory records where an optimizer has been; it need not exhaust the capability obtainable from the updates it has computed.

\begin{figure}[tbp]
 \centering
 \includegraphics[width=\linewidth]{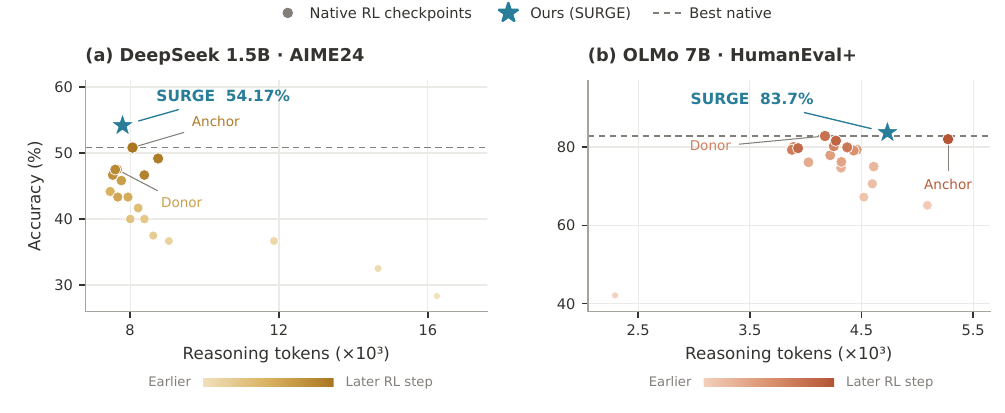}
 \caption{\textbf{Scaling beyond observed RL checkpoints, from 1.5B math to 7B code.} Circles encode native RL step (color) and accuracy (size); SURGE exceeds the best measured native accuracy (dashed line) with fewer tokens than the anchor. Each panel shows 19 checkpoints on linear axes. Token counts are in thousands.}
 \label{fig:trajectory}
\end{figure}

This distinction changes what counts as the output of RL. Checkpoint selection treats a run as a finite list of complete policies. Viewed in parameter space, the same history also contains distinct updates relative to a common initialization. Those updates can support policies that the optimizer never visited. Because reasoning performance is nonlinear in the weights, recombining update components need not average the capabilities of their sources. The relevant question is therefore whether stored history supports a higher accuracy ceiling than native checkpoint selection alone.

We study \emph{policy-space scaling}: expanding the deployable policy set accessible from a fixed RL history. Training-time scaling expands what the optimizer can learn by extending optimization. Test-time scaling expands what a fixed policy can compute by extending inference. Policy-space scaling expands what a completed training history can deploy by constructing additional policies from already-computed updates. Offline policy construction and evaluation have a cost, but the resulting model has the same architecture and runs as one policy at deployment.

Our contributions are threefold:
\begin{itemize}
\setlength{\itemsep}{2pt}
\setlength{\parsep}{0pt}
\item We show that an RL history can support policies stronger than its best measured checkpoint, and formulate \emph{policy-space scaling}: expanding the deployable policy set from already-computed updates, without further RL or a larger inference budget.
\item We develop \surge\ to instantiate this axis through delta-first spectral construction, preserving an anchor's protected component and admitting a donor's complementary component. Given a source pair and an empirical energy target, weight-only calibration determines the setting without an accuracy sweep.
\item We demonstrate this opportunity across 1.5B math and 7B code. SURGE exceeds the best of 19 measured native checkpoints on DeepSeek AIME24 (54.17\% vs.\ 50.83\%) and OLMo HumanEval+ (83.7\% vs.\ 82.8\%; Figure~\ref{fig:trajectory}). Across nine math benchmarks, average accuracy improves over the anchors by 2.08/1.78 points for DeepSeek/Nemotron, with fewer tokens on every benchmark. Geometric controls support the importance of RL-update structure.
\end{itemize}

\section{Methodology}
\label{sec:method}

\subsection{An RL trajectory as a resource for scaling}
\label{sec:trajectory}
Let $\theta_0$ initialize RL and $\mathcal C_T=\{\theta_t:t\in\mathcal I_T\}$ contain its stored checkpoints with nonzero RL updates through step $T$. Native checkpoint selection is restricted to $\mathcal C_T$. A construction $F$ acting on two compatible checkpoints opens a larger policy family,
\begin{equation}
 \mathcal P_T=\mathcal C_T\cup
 \{F(\theta_a,\theta_b;f):a,b\in\mathcal I_T,\ a\ne b,\ f\in\mathcal F\}.
 \label{eq:policy-family}
\end{equation}
Training-time scaling extends the history; policy-space scaling explores $\mathcal P_T$ while holding that history fixed. Equation~\ref{eq:policy-family} defines a candidate space, not a requirement to enumerate all pairs. We select an accuracy-strong checkpoint as the \emph{anchor} and a competitive, shorter-response checkpoint as the \emph{donor}, using source-policy statistics already available along the RL trajectory. These roles need not follow a temporal ordering or a per-benchmark accuracy ranking. Source selection is distinct from the weight-only FC calibration in Section~\ref{sec:calibration}.

For sources at steps $t_a,t_b$, set $T=\max(t_a,t_b)$ and let $A(\theta)$ denote measured accuracy under a fixed protocol. The question is whether the constructed policy exceeds $\max_{\theta\in\mathcal C_T}A(\theta)$ without further RL or a larger inference budget. When the later source is the last evaluated checkpoint, this comparison covers the full measured history. Reasoning tokens count the full generated response, including submitted code. We study improvements attainable from a fixed history, without claiming a fitted scaling law or improvement for every pair.

\subsection{Delta-first singular decomposition}
\begin{figure}[htbp]
 \centering
 \includegraphics[width=\linewidth]{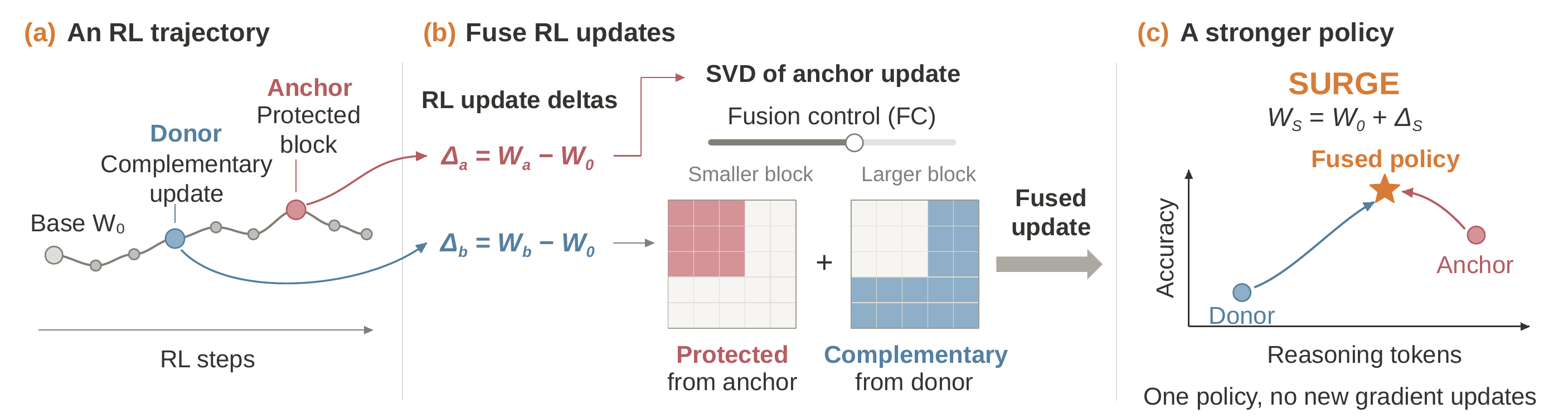}
 \caption{\textbf{SURGE overview.} Checkpoints from one RL trajectory are expressed as deltas before SVD. The anchor defines the protected update block; the donor supplies its complement. Recombination yields one policy without new gradient updates. The accuracy--token plot is schematic.}
 \label{fig:overview}
\end{figure}

Figure~\ref{fig:overview} summarizes the construction. The anchor $\theta_a$ and donor $\theta_b$ share an architecture and RL initialization. The method is asymmetric: the anchor defines the protected block, so exchanging the inputs generally changes the output.

For each target matrix $W\in\mathbb{R}^{m\times n}$, we first form the updates relative to the start of RL:
\begin{equation}
  \Delta_a=W_a-W_0,\qquad \Delta_b=W_b-W_0.
  \label{eq:delta}
\end{equation}
Only after this subtraction do we compute the economy singular value decomposition
\begin{equation}
  \Delta_a=U\Sigma V^\top,\qquad r=\min(m,n),\qquad
  k=\lceil \FC\,r\rceil,\quad 0<\FC\leq 1.
  \label{eq:svd}
\end{equation}
The left and right singular vectors are eigenvectors of $\Delta_a\Delta_a^\top$ and $\Delta_a^\top\Delta_a$, respectively. Here \emph{fusion control} (FC) sets the fraction of singular directions used to define a protected block. Applying SVD to $W_a$ instead would define a different geometry, dominated in part by the shared initialization. Delta-first construction makes the reference the change induced by RL itself. A component shared by the initialization and both checkpoints cancels before the spectral basis is computed. The resulting coordinates describe accumulated RL changes, rather than the absolute weight structure inherited before RL. Although $W_0$ cancels from the checkpoint difference $W_b-W_a$, it remains essential to this geometry.

\subsection{Protected and complementary update components}
Let $U_k,V_k$ contain the leading $k$ left and right singular vectors of $\Delta_a$. In matrix space with the Frobenius inner product, define
\begin{align}
 \mathcal B_{a,k}&=\{U_k Z V_k^\top:Z\in\mathbb R^{k\times k}\},\nonumber\\
 \Keep_{a,k}(X)&=(U_kU_k^\top)X(V_kV_k^\top),
 &\Comp_{a,k}(X)&=X-\Keep_{a,k}(X).
 \label{eq:projector}
\end{align}
We call $\mathcal B_{a,k}$ the \emph{protected update block}. The \emph{protected-block projector} $\Keep_{a,k}$ extracts the retained component, while the \emph{complement projector} $\Comp_{a,k}$ extracts the update outside that block. The index $a$ emphasizes that both operators are defined by the anchor's RL update. The protected object is a two-sided matrix block of dimension $k^2$, including interactions between the selected left and right directions, rather than only $k$ matched singular outer products.

SURGE preserves the anchor's protected component and admits the donor's complementary component:
\begin{align}
 \Delta_{\mathrm S}&=\Delta_b+\Keep_{a,k}(\Delta_a-\Delta_b),\nonumber\\
 W_{\mathrm S}&=W_0+\Delta_{\mathrm S}.
 \label{eq:fusion}
\end{align}
The construction gives exact component identities,
$\Keep_{a,k}(\Delta_{\mathrm S})=\Keep_{a,k}(\Delta_a)$ and
$\Comp_{a,k}(\Delta_{\mathrm S})=\Comp_{a,k}(\Delta_b)$.
These are statements about parameters. Because model computation is nonlinear, they do not imply independent preservation of the source policies' behaviors.

\subsection{Calibrating fusion control by retained update energy}
\label{sec:calibration}
FC controls the protected block's size, not an interpolation amplitude. To relate this rank fraction to the anchor's update geometry, define the retained energy
\begin{equation}
 \xi_a(\FC)=\frac{\sum_{\ell\in\mathcal L}\sum_{i=1}^{k_\ell}\sigma_{\ell,i}^2}
 {\sum_{\ell\in\mathcal L}\sum_{i=1}^{r_\ell}\sigma_{\ell,i}^2},
 \qquad k_\ell=\lceil\FC r_\ell\rceil,
 \label{eq:energy}
\end{equation}
where $\mathcal L$ contains the target matrices. This is the fraction of anchor-update squared Frobenius norm preserved by the protected blocks. It depends on the anchor, not the donor. The entire curve follows from cumulative squared singular values of the same SVD, without additional decompositions or inference. FC is the rank coordinate used by the algorithm; $\xi_a$ calibrates what that coordinate preserves in the anchor's update. We use the discrete rule
\begin{equation}
 \FC_{\mathrm{spec}}=\arg\min_{f\in\mathcal F}|\xi_a(f)-\tau|,
 \qquad \tau=0.95,\quad \mathcal F=\{0.1,0.2,\ldots,0.9\}.
 \label{eq:fc-selection}
\end{equation}
We identified $\tau=0.95$ from the initial mathematical study and fixed it before applying the same procedure to OLMo. The common rule returns FC $=0.8$, $0.8$, and $0.7$ for DeepSeek, Nemotron, and OLMo, respectively. Once the checkpoint pair and target are fixed, no accuracy sweep is required: cumulative singular values determine FC, and one fusion produces the policy. The target is empirical, not a guarantee of reward optimality; the rule selects the nearest candidate without requiring $\xi_a\geq\tau$. Section~\ref{sec:analysis} analyzes the resulting policy family.

For a fixed projector, fusion is the closest donor update whose protected component matches the anchor. Appendix~\ref{app:implementation} gives the variational characterization, proof, and details of the native references.

\subsection{Why an optimization path need not exhaust capability}
RL optimizes a coupled policy and realizes one sequence of joint parameter updates. It does not enumerate the recombinations of updates accumulated along that sequence. These are distinct spaces: the optimizer's visited checkpoints and the policies constructible from their components. Since policy behavior is nonlinear in the weights, scores of the latter need not lie between those of their sources. Section~\ref{sec:experiments} tests this possibility against native trajectories and geometric controls.

A shorter-response donor may avoid some unproductive exploration or redundant continuation. If its useful updates complement the anchor, fusion can improve correctness as well as efficiency; the ablations test whether structured recombination matters beyond reducing token usage.

When both $\Keep_{a,k}(W_b-W_a)$ and $\Comp_{a,k}(W_b-W_a)$ are nonzero, the fusion in Equation~\ref{eq:fusion} lies off the affine line through the two source matrices. It explores a recombination that no scalar interpolation coefficient can reproduce. We use ``beyond the RL trajectory'' in the performance sense: exceeding the measured native policies. Appendix~\ref{app:implementation} gives a conditional local reward interpretation.

\paragraph{Implementation.}
We fuse the query, key, value, and output attention projections and the gate, up, and down feed-forward projections, using one FC across layers. Embeddings, normalization parameters, language-model heads, and attention biases are copied from the anchor. Algorithm~\ref{alg:surge} states the tensor-level procedure. SVD and fusion arithmetic use FP32, and checkpoints are saved in BF16. The output runs as one policy. For the two math families, a full pass over 196 matrices, including spectral diagnostics and checkpoint writing, takes 37.2--40.1 seconds on one RTX A6000 with warm caches (Appendix~\ref{app:compute}).

\section{Experiments}
\label{sec:experiments}
We evaluate whether a fixed RL history supports stronger policies than its measured checkpoints, whether this opportunity persists across scale and domain, and which update structure matters. Main tables evaluate one pair from each of three histories. Full-trajectory comparisons cover 19 native checkpoints each on DeepSeek AIME24 and OLMo HumanEval+, through their respective anchors.
\subsection{Experimental setup}
\paragraph{RL training follows JustRL.}
The two math trajectories use DeepSeek-R1-Distill-Qwen-1.5B and OpenMath-Nemotron-1.5B, following JustRL's single-stage GRPO recipe \citep{he2025justrl,deepseek2025r1,moshkov2025openmath}. Training uses veRL, DAPO-Math-17k, and binary correctness rewards from the DAPO rule-based verifier \citep{shao2024deepseekmath,sheng2025hybridflow,yu2025dapo}. Each prompt produces eight rollouts at temperature 1.0; group-relative rewards drive updates with learning rate $10^{-6}$ and batch size 256. The recipe uses no explicit length penalty, KL loss, entropy regularization, or stage changes. Appendix~\ref{app:implementation} provides the full settings.

\paragraph{Benchmarks and evaluation.}
We use JustRL's nine mathematical benchmarks: AIME24/25, HMMT25, BRUMO25, CMIMC25, AMC23, MATH-500 (MATH), Minerva, and OlympiadBench (Olympiad) \citep{li2024numinamath,balunovic2025matharena,hendrycks2021math,lewkowycz2022minerva,he2024olympiadbench}. Accuracy averages correctness over four responses per problem; reasoning tokens are averaged over generated responses. Decoding follows JustRL with temperature 0.7, top-$p$ 0.9, and a 32K response limit. Scoring combines rule-based evaluation and CompassVerifier \citep{he2025justrl}. These are sampled single-response accuracies, not best-of-four success rates. Avg. in Table~\ref{tab:main} weights benchmarks equally for both accuracy and tokens. Appendix~\ref{app:variability} reports DeepSeek accuracy variation across the four evaluation rounds.

\paragraph{Evaluated configurations.}
The math experiments use DeepSeek steps 3600/2600 and Nemotron steps 3440/1200 as anchor/donor, respectively. The anchors have strong aggregate accuracy; the donors offer shorter responses while retaining competitive accuracy in the available source-policy evaluations. Equation~\ref{eq:fc-selection} returns FC $=0.8$ for both. Each comparison holds the RL history, architecture, and decoding protocol fixed.

\begin{table}[htbp]
\caption{\textbf{Main results.} Accuracy (\%) and mean reasoning tokens occupy separate labeled rows. Bold marks the best accuracy per family and benchmark. Avg. is the unweighted mean across benchmarks. Each $\Delta$ compares SURGE with its anchor: accuracy in percentage points and tokens as relative change; green indicates improvement and red indicates regression. Changes are computed before rounding.}
\label{tab:main}
\centering
\fontsize{8.5}{10.2}\selectfont
\setlength{\tabcolsep}{1.35pt}
\renewcommand{\arraystretch}{1.12}
\begin{tabular*}{\linewidth}{@{\extracolsep{\fill}}ll!{\color{surgegrid}\vrule width .25pt}c*{9}{!{\color{surgegrid}\vrule width .25pt}c}@{}}
\toprule
Policy & Metric & AIME24 & AIME25 & HMMT25 & BRUMO25 & CMIMC25 & AMC23 & MATH & Minerva & Olympiad & Avg. \\
\midrule
\multicolumn{12}{@{}l}{\textbf{DeepSeek}} \\[2pt]
\multirow{2}{*}{Donor} & Acc. & 47.50 & 35.83 & 19.17 & 46.67 & 21.88 & 88.13 & 90.75 & 50.64 & 64.80 & 51.71 \\
 & Tokens & 7,651 & 6,912 & 7,574 & 7,194 & 8,525 & 4,753 & 3,115 & 4,582 & 5,570 & 6,208 \\[3pt]
\multirow{2}{*}{Anchor} & Acc. & 50.83 & 36.67 & 20.00 & 49.17 & 22.50 & 90.00 & 91.60 & 52.02 & 66.58 & 53.26 \\
 & Tokens & 8,065 & 7,927 & 8,109 & 7,892 & 9,147 & 5,472 & 3,656 & 4,796 & 6,246 & 6,812 \\[3pt]
\multirow{2}{*}{\surge} & Acc. & \textbf{54.17} & \textbf{38.33} & \textbf{23.33} & \textbf{51.67} & \textbf{25.63} & \textbf{92.50} & \textbf{92.40} & \textbf{52.76} & \textbf{67.32} & \textbf{55.35} \\
 & Tokens & 7,798 & 7,383 & 7,693 & 7,285 & 8,700 & 5,225 & 3,415 & 4,742 & 5,832 & 6,453 \\[3pt]
\cmidrule(lr){1-12}
\multirow{2}{*}{$\Delta$} & Acc. & \textcolor{surgegreen}{+3.34} & \textcolor{surgegreen}{+1.66} & \textcolor{surgegreen}{+3.33} & \textcolor{surgegreen}{+2.50} & \textcolor{surgegreen}{+3.13} & \textcolor{surgegreen}{+2.50} & \textcolor{surgegreen}{+0.80} & \textcolor{surgegreen}{+0.74} & \textcolor{surgegreen}{+0.74} & \textcolor{surgegreen}{+2.08} \\
 & Tokens & \textcolor{surgegreen}{-3.3\%} & \textcolor{surgegreen}{-6.9\%} & \textcolor{surgegreen}{-5.1\%} & \textcolor{surgegreen}{-7.7\%} & \textcolor{surgegreen}{-4.9\%} & \textcolor{surgegreen}{-4.5\%} & \textcolor{surgegreen}{-6.6\%} & \textcolor{surgegreen}{-1.1\%} & \textcolor{surgegreen}{-6.6\%} & \textcolor{surgegreen}{-5.3\%} \\
\midrule
\multicolumn{12}{@{}l}{\textbf{Nemotron}} \\[2pt]
\multirow{2}{*}{Donor} & Acc. & 60.00 & 50.83 & 35.00 & 61.67 & 29.38 & 91.88 & 92.45 & 33.18 & 73.81 & 58.69 \\
 & Tokens & 9,212 & 10,387 & 10,652 & 8,844 & 10,847 & 5,687 & 3,754 & 6,142 & 6,536 & 8,007 \\[3pt]
\multirow{2}{*}{Anchor} & Acc. & 68.33 & 60.83 & 35.83 & 65.83 & 38.75 & 95.00 & 94.25 & 33.55 & \textbf{77.26} & 63.29 \\
 & Tokens & 13,030 & 14,395 & 14,841 & 12,548 & 15,603 & 8,393 & 5,452 & 8,627 & 9,396 & 11,365 \\[3pt]
\multirow{2}{*}{\surge} & Acc. & \textbf{70.00} & \textbf{63.33} & \textbf{39.17} & \textbf{68.33} & \textbf{41.88} & \textbf{97.50} & \textbf{94.35} & \textbf{34.28} & 76.85 & \textbf{65.08} \\
 & Tokens & 11,395 & 12,294 & 13,071 & 11,061 & 13,223 & 7,375 & 4,911 & 7,513 & 8,348 & 9,910 \\[3pt]
\cmidrule(lr){1-12}
\multirow{2}{*}{$\Delta$} & Acc. & \textcolor{surgegreen}{+1.67} & \textcolor{surgegreen}{+2.50} & \textcolor{surgegreen}{+3.34} & \textcolor{surgegreen}{+2.50} & \textcolor{surgegreen}{+3.13} & \textcolor{surgegreen}{+2.50} & \textcolor{surgegreen}{+0.10} & \textcolor{surgegreen}{+0.73} & \textcolor{surgered}{-0.41} & \textcolor{surgegreen}{+1.78} \\
 & Tokens & \textcolor{surgegreen}{-12.5\%} & \textcolor{surgegreen}{-14.6\%} & \textcolor{surgegreen}{-11.9\%} & \textcolor{surgegreen}{-11.9\%} & \textcolor{surgegreen}{-15.3\%} & \textcolor{surgegreen}{-12.1\%} & \textcolor{surgegreen}{-9.9\%} & \textcolor{surgegreen}{-12.9\%} & \textcolor{surgegreen}{-11.2\%} & \textcolor{surgegreen}{-12.8\%} \\
\bottomrule
\end{tabular*}
\end{table}

\subsection{Scaling a fixed mathematical-reasoning history}
\label{sec:results}
SURGE raises DeepSeek's measured AIME24 accuracy ceiling from 50.83\% to 54.17\% at the same RL horizon (Figure~\ref{fig:trajectory}). It exceeds all 19 native measurements while using 7,798 tokens versus the anchor's 8,065. The benefit extends across the evaluation suite: accuracy improves and reasoning tokens fall on all nine benchmarks. Avg. accuracy rises from 53.26\% to 55.35\% (+2.08 points), with 5.3\% fewer tokens.

Nemotron's Avg. rises from 63.29\% to 65.08\% (+1.78 points), with 12.8\% fewer tokens. Accuracy improves on eight of nine benchmarks; only Olympiad decreases, by 0.41 points. Reasoning tokens fall on all nine. AIME24 reaches 70.00\%, above the anchor's 68.33\% and donor's 60.00\%, with 12.5\% fewer tokens than the anchor.

\begin{figure}[htbp]
 \centering
 \includegraphics[width=\linewidth]{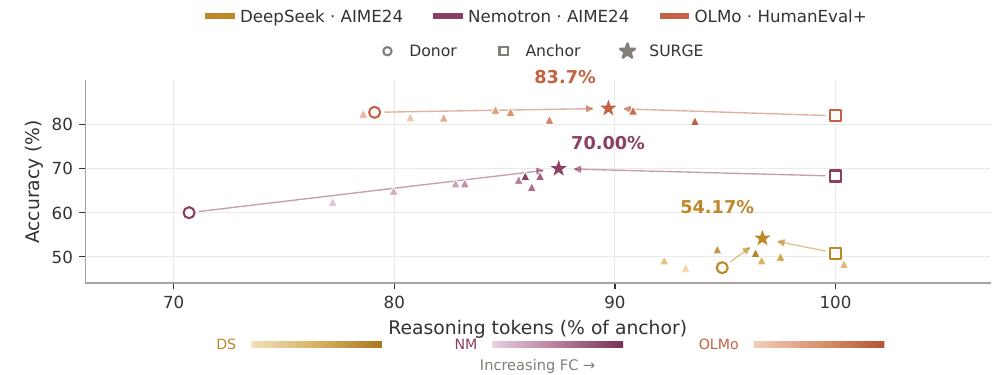}
 \caption{\textbf{New policies from fixed RL histories.} Tokens are normalized to each anchor. Triangles are measured FC candidates; stars mark the spectrally calibrated policies. Arrows connect source checkpoints to the resulting policy. All candidates use four rounds; comparisons are within each family and benchmark.}
 \label{fig:tradeoff}
\end{figure}

Across all three histories in Figure~\ref{fig:tradeoff}, the spectrally calibrated policies exceed both sources on the displayed benchmark, with fewer tokens than the anchor under the same response budget.

\subsection{Cross-scale and cross-domain validation: 7B code}
\label{sec:coding}
Olmo-3.1-7B-RL-Zero-Code changes the scale, architecture, and reasoning domain. Training starts from Olmo-3-1025-7B and follows OlmoRL on Dolci-RL-Zero-Code-7B: GRPO-based execution rewards, eight responses per prompt, learning rate $10^{-6}$, and a 16K response limit \citep{olmo2025olmo3}. HumanEval+ and MBPP+ use EvalPlus \citep{liu2023evalplus}, with four rounds of one response per problem at temperature 1.0, top-$p$ 1.0, and a 32K limit. Appendix~\ref{app:coding} gives the full protocol and 19 native checkpoint measurements.

Step 1950 has the highest two-benchmark mean among six jointly evaluated checkpoints and serves as the anchor. The shorter-response donor is step 1500. Equation~\ref{eq:fc-selection} selects FC $=0.7$ from the anchor spectrum with the fixed target.

\begin{figure}[tbp]
 \centering
 \begin{minipage}[t]{0.495\linewidth}
  \vspace{0pt}\centering
  \includegraphics[width=\linewidth]{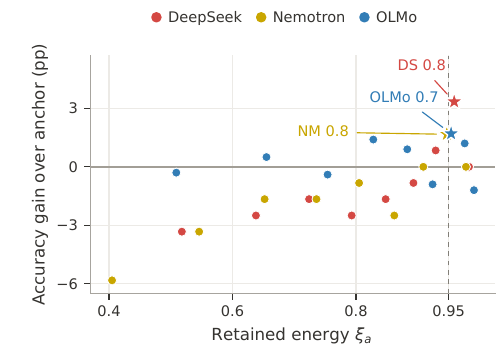}
  \caption{\textbf{Shared spectral calibration across trajectories.} Measured candidates (dots) and selected FC settings (stars); dashed line: $\xi_a=0.95$. AIME24 for math, HumanEval+ for code.}
  \label{fig:fc}
 \end{minipage}\hfill
 \begin{minipage}[t]{0.475\linewidth}
  \vspace{0pt}
  \centering\small
\captionof{table}{\textbf{7B coding results.} Accuracy (\%) and reasoning tokens (k). $\Delta$ gives changes from the anchor in points and percent.}
\label{tab:coding}
\setlength{\tabcolsep}{2.5pt}
\renewcommand{\arraystretch}{1.28}
\begin{tabular}{@{}ll!{\color{surgegrid}\vrule width .35pt}c!{\color{surgegrid}\vrule width .35pt}c@{}}
\toprule
Policy & Metric & HumanEval+ & MBPP+\\
\midrule
Donor & Acc. (\%) & 82.8 & 68.4\\
 & Tokens (k) & 4.17 & 3.02\\
\midrule
Anchor & Acc. (\%) & 82.0 & \textbf{71.2}\\
 & Tokens (k) & 5.28 & 3.86\\
\midrule
\textbf{SURGE} & Acc. (\%) & \textbf{83.7} & \textbf{71.2}\\
 & Tokens (k) & 4.73 & 3.45\\
\midrule
$\Delta$ & Accuracy & \textcolor{surgegreen}{+1.7} & \textcolor{surgegreen}{0.0}\\
 & Tokens & \textcolor{surgegreen}{$-10.3\%$} & \textcolor{surgegreen}{$-10.5\%$}\\
\bottomrule
\end{tabular}

 \end{minipage}
\end{figure}

Table~\ref{tab:coding} reports 83.7\% HumanEval+ accuracy: 1.7 points above the anchor and 0.9 above the best native checkpoint, the donor. MBPP+ matches the anchor at reported precision. Tokens fall by 10.3\% and 10.5\% relative to the anchor. These point estimates extend the observed gain beyond native checkpoint selection to 7B code.

\paragraph{A concrete policy difference.}
Table~\ref{tab:case-code-chars} shows a case where SURGE more consistently follows the program specification. Direct set equality appears in three of its four responses, versus one for the donor and none for the anchor. This selected example illustrates a behavioral difference, rather than estimating its prevalence; Appendix~\ref{app:cases} provides additional cases.
\begin{table}[tbp]
\caption{\textbf{Specification fidelity on HumanEval+ \#54.} A selected OLMo coding example. Correctness counts are over four responses per policy, each checked against the original and additional EvalPlus tests.}
\label{tab:case-code-chars}
\centering\small
\setlength{\tabcolsep}{6pt}
\renewcommand{\arraystretch}{1.10}
\begin{tabularx}{\linewidth}{@{}Y!{\color{surgegrid}\vrule width .35pt}Y!{\color{surgegrid}\vrule width .35pt}Y@{}}
\toprule
\multicolumn{3}{@{}p{\linewidth}@{}}{\textbf{Task.} Implement \texttt{same\_chars(s0, s1)}: compare the two strings' character sets. Order and repetition do not matter; case matters.}\\[3pt]
\midrule
\textbf{Donor} & \textbf{Anchor} & \textbf{SURGE}\\
\midrule
One response compares sets correctly. \textcolor{surgered}{The others compare sorted characters, lowercase the strings, or compare only set sizes.}
& \textcolor{surgered}{Two responses count character multiplicities; two compare sets after lowercasing. These reject valid repetitions or erase case distinctions.}
& \textcolor{surgeteal}{Three responses compare sets directly, ignoring repetition while preserving case.} One response still sorts lowercased strings.\\[5pt]
Observed correct: \textbf{1/4} & Observed correct: \textbf{0/4} & Observed correct: \textbf{3/4}\\[3pt]
\midrule
\multicolumn{3}{@{}l@{}}{\textbf{Passing implementation:}\quad \texttt{return set(s0) == set(s1)}}\\[2pt]
\bottomrule
\end{tabularx}
\end{table}

\subsection{A shared spectral coordinate across trajectories}
\label{sec:analysis}

Figure~\ref{fig:fc} compares accuracy gains in retained-energy space. The fixed-target, weight-only rule selects strong policies across all three trajectories despite different rank fractions. Matching retained energy accounts for differences in matrix shape and update spectra that raw FC misses (Appendix~\ref{app:spectral}). The sweeps characterize sensitivity; once the target is fixed, new pairs require no accuracy sweep.

The calibrated settings coincide with the highest measured accuracies in all three sweeps. On HumanEval+, the neighboring FC $=0.8$ reaches 83.2\%, versus 83.7\% for the rule's FC $=0.7$; these small differences do not establish a statistically resolved optimum. Other useful policies exist in the same history: FC $=0.4$ reaches 83.4\% with 4,462 tokens.

Accuracy is non-monotonic even though retained energy increases with FC: parameter-space motion does not linearly interpolate policy quality. DeepSeek at FC $=0.7$ reaches 51.67\% with 7,633 tokens; $0.8$ adds 2.50 points for 165 tokens. This family supports different deployment choices; spectral calibration selects one policy without evaluating every candidate (raw-FC measurements: Appendix~\ref{app:sweeps}).

\subsection{Ablations: which update geometry matters?}
\label{sec:ablation}
\begin{table}[htbp]
\caption{\textbf{DeepSeek checkpoint-reuse controls on AIME24.} Four evaluation rounds per policy. Scalar and spectral controls use the same source pair; late averaging uses its RL history. The random row averages four constructions. Full scalar and averaging results are in Appendix~\ref{app:ablations}.}
\label{tab:ablation}
\centering\small
\begin{tabular*}{\linewidth}{@{\extracolsep{\fill}}ccc@{}}
\toprule
Method & Accuracy (\%) & Reasoning tokens \\
\midrule
Anchor & 50.83 & 8,065 \\
Donor & 47.50 & 7,651 \\
Best tested non-endpoint scalar ($\alpha=-0.5$) & 49.17 & 9,141 \\
Linear interpolation ($\alpha=0.71$) & 47.50 & 7,659 \\
Late checkpoint average (3 checkpoints) & 50.83 & 8,078 \\
Anchor-only spectral truncation & 44.17 & 8,320 \\
Absolute-weight SVD & 41.67 & 8,173 \\
Random subspace (mean of 4) & 47.29 & 7,574 \\
Trailing singular subspace & 44.17 & 7,743 \\
\midrule
\surge\ (Delta-SVD) & \textbf{54.17} & 7,798 \\
\bottomrule
\end{tabular*}
\end{table}

Table~\ref{tab:ablation} compares reuse of the same DeepSeek history. Scalar and spectral controls act on 196 body linear matrices; checkpoint averages include all parameters.

Replacing delta-first SVD with absolute-weight SVD reduces AIME24 accuracy from 54.17\% to 41.67\%, with lower accuracy on all nine benchmarks (Appendix~\ref{app:ablations}). At the same FC and block size, the absolute-weight block captures 28.5\% of RL-update energy, versus 95.9\% for the delta-defined block (Appendix~\ref{app:spectral}).

\paragraph{The donor contributes beyond anchor truncation.}
Retaining only the anchor's protected update, $W_{\mathrm{trunc}}=W_0+\Keep_{a,k}(\Delta_a)$, uses the same projector and FC $=0.8$. It reaches 44.17\% accuracy with 8,320 tokens, versus SURGE's 54.17\% and 7,798. Removing the anchor tail alone does not recover the gain: adding the donor's complementary update raises measured accuracy by 10.00 points in this pair.

\paragraph{Scalar combinations and checkpoint averaging.}
For $W_{\alpha}=(1-\alpha)W_a+\alpha W_b$, six tested interpolation and extrapolation coefficients reach at most 49.17\% ($\alpha=-0.5$), below the anchor and SURGE. To control for displacement magnitude, define
\begin{equation}
 \eta=\frac{\left(\sum_{\ell\in\mathcal L}\|W_{\mathrm S}^{\ell}-W_a^{\ell}\|_F^2\right)^{1/2}}
 {\left(\sum_{\ell\in\mathcal L}\|W_b^{\ell}-W_a^{\ell}\|_F^2\right)^{1/2}}
 =0.713,
 \label{eq:displacement}
\end{equation}
Scalar combinations have normalized distance $|\alpha|$. Thus $\alpha=0.71$ approximately matches SURGE's aggregate distance, yet reaches only 47.50\%; direction matters beyond magnitude. Three- and six-checkpoint late averages reach 50.83\% and 49.17\%, respectively. Neither tested average exceeds the anchor. Appendix~\ref{app:ablations} gives the full grid and averaging windows.

\paragraph{Protected geometry beyond token reduction.}
At fixed block size, random left/right orthonormal bases yield 47.29\% mean accuracy across four independent constructions. Retaining the last $k$ singular directions instead yields 44.17\%. Both use fewer tokens than SURGE (7,574 and 7,743 versus 7,798), while losing 6.88 and 10.00 accuracy points. These controls favor the leading RL-update block over the tested alternatives; neither block size nor shorter reasoning alone recovers the gain.

Across the repeated math evaluations, within-cell accuracy and reasoning tokens have a weak negative association ($r=-0.181$, permutation $p=0.021$), consistent with but not establishing a causal efficiency benefit (Appendix~\ref{app:round-correlation}).

\paragraph{Implications for policy-space scaling.}
The controls distinguish expanding the deployable policy set from merely perturbing weights or shortening responses. This expansion spends offline construction compute while reusing the completed RL history. Once a checkpoint pair and retention target are fixed, the anchor spectrum determines one policy; evaluating the broader family is optional and has a separate cost. The reported sweeps diagnose that family, whereas deployment uses a single checkpoint under the original decoding protocol. Thus, the additional scaling opportunity lies in how already-computed RL updates are used.

\section{Related Work}
\paragraph{Scaling reasoning with verifiable rewards.}
DeepSeekMath introduced GRPO for mathematical reasoning, and DeepSeek-R1 demonstrated that large-scale RL can elicit effective reasoning behavior from correctness rewards \citep{shao2024deepseekmath,deepseek2025r1}. DAPO develops an open RL system with dynamic sampling and changes to clipping and token-level optimization \citep{yu2025dapo}. ProRL studies prolonged training, using KL control and reference-policy resets to sustain exploration and expand reasoning performance \citep{liu2025prorl}. JustRL emphasizes a fixed, single-stage recipe for small reasoning models \citep{he2025justrl}; we follow its setting and study how to reuse its accumulated updates. Efficiency is also a property of the training objective: Dr.\ GRPO identifies length-related bias in GRPO and shows that correcting it can improve both performance and token efficiency \citep{liu2025drgrpo}. At inference time, s1 controls reasoning compute through budget forcing \citep{muennighoff2025s1}. SURGE complements these training- and inference-time approaches by constructing one deployable policy from stored checkpoints, without further policy-gradient updates or a larger decoding budget.

\paragraph{Model merging and update composition.}
Model soups average fine-tuned weights without inference-time ensembling, while Fisher-weighted averaging accounts for parameter importance \citep{wortsman2022soups,matena2021fisher}. Task arithmetic composes updates relative to a common initialization; TIES-Merging reduces sign interference, and DARE sparsifies and rescales updates \citep{ilharco2023task,yadav2023ties,yu2024dare}. Spectral methods address their matrix structure: Task Singular Vectors reduces interference among task-update directions, and isotropic merging rebalances spectra using common and task-specific subspaces \citep{gargiulo2024tsv,marczak2025isotropic}. More directly, ResMerge finds useful information in both the leading heads and residuals of RL task vectors, combining multiple RL experts through residual consensus and agreement-gated head correction \citep{sun2026resmerge}. These studies establish tools for update composition. Our central question concerns the capability accessible from one RL training history. SURGE uses spectral composition to expand that history's candidate policy space, and evaluates whether it exceeds the measured native trajectory. These experiments examine policy-space scaling from already-computed RL updates across domains and through geometric controls.

\paragraph{Reusing RL training trajectories.}
AlphaRL predicts dominant update components from early checkpoints through accuracy-conditioned regression \citep{cai2025predictability}. RELEX decomposes stacked checkpoint deltas and linearly extrapolates their rank-1 trajectory coefficients \citep{wei2026relex}. NExt trains a nonlinear predictor of rank-1 update representations from LoRA-based RLVR trajectories \citep{chen2026next}. These methods forecast additional policies from observed training history. Extrapolative weight averaging also extends correctness--efficiency frontiers in code RL, combining policies trained under different unit-test coverage \citep{zheng2026ewa}. There, efficiency concerns generated programs under execution resource limits; here, we measure reasoning tokens. SURGE expands the deployable policy set by recombining components of observed RL updates, with a weight-only calibration rule after source selection. It uses the history's update geometry without fitting a trajectory predictor or specifying a future training step. This offers a complementary construction for policy-space scaling from an existing RL budget.

\section{Conclusion}
An RL run can deliver more capability than its best measured checkpoint. Across 1.5B math and 7B code, one weight-only calibration procedure constructs policies that improve on their sources; on the evaluated DeepSeek and OLMo trajectories, they exceed every measured native checkpoint. The optimizer's path therefore need not define the capability ceiling of its training history. Policy-space scaling turns that history into a reusable policy space, complementing new optimization and additional inference with offline reuse of already-computed updates. Its construction and evaluation costs are explicit, while deployment remains a single policy. A broader RL scaling strategy should account both for how updates are produced and for how their accumulated history is used: additional capability can reside in policies the optimizer never visited.

\FloatBarrier
\begingroup
\interlinepenalty=10000
\bibliographystyle{unsrtnat}
\bibliography{main}
\endgroup

\clearpage
\appendix
\fontsize{10}{12}\selectfont
\raggedbottom
\section{Training, evaluation, and fusion details}
\label{app:implementation}
Table~\ref{tab:settings} records the math training and evaluation settings; Appendix~\ref{app:coding} gives the code settings. All main-table results use four responses per question, unlike JustRL's higher sample count on several tasks. The qualitative examples in Appendix~\ref{app:cases} are per-problem diagnostics.

\begin{table}[H]
\caption{\textbf{Math configuration.} Training settings follow JustRL; evaluation settings and four-response main-table estimates are shared by each fusion and its source checkpoints.}
\label{tab:settings}
\centering\small
\begin{tabular}{@{}lll@{}}
\toprule
Stage & Setting & Value\\
\midrule
RL & Algorithm / framework & GRPO / veRL\\
& Training data & DAPO-Math-17k\\
& Reward & Binary correctness, DAPO rule-based verifier\\
& Learning rate & $10^{-6}$, constant\\
& Train batch / PPO mini-batch & 256 / 64\\
& Micro-batch per GPU & 1\\
& Responses per prompt / temperature & 8 / 1.0\\
& Clip-ratio interval & $[0.8,1.28]$\\
& Maximum prompt / response & 1K / 15K tokens\\
& KL loss / entropy regularization & None / none\\
& Length penalty / stage schedule & None / single stage\\
\midrule
Evaluation & Temperature / top-$p$ & 0.7 / 0.9\\
& Maximum response & 32K tokens\\
& Main-table responses per problem & 4 for every benchmark\\
& Verification & Rule-based scoring plus CompassVerifier\\
\midrule
Fusion & Primary FC & 0.8 for both math families ($\xi_a$ nearest 0.95)\\
& Reference for SVD & Anchor update $W_a-W_0$\\
& Computation / saved precision & FP32 / BF16\\
& Non-target parameters & Copied from anchor\\
\bottomrule
\end{tabular}
\end{table}

\begin{algorithm}[H]
\caption{SURGE: weight-only calibration and policy construction}
\label{alg:surge}
\small
\begin{algorithmic}[1]
\Require Initialization $\theta_0$, anchor $\theta_a$, donor $\theta_b$, target matrices $\mathcal L$, grid $\mathcal F$ and target $\tau$ from Equation~\ref{eq:fc-selection}
\For{each matrix $\ell\in\mathcal L$}
 \State $\Delta_a^\ell\gets W_a^\ell-W_0^\ell$, \quad $\Delta_b^\ell\gets W_b^\ell-W_0^\ell$
 \State Cache $U^\ell,\Sigma^\ell,(V^\ell)^\top\gets\operatorname{SVD}(\Delta_a^\ell)$
\EndFor
\State Compute $\xi_a(f)$ for $f\in\mathcal F$ from cached singular values (Equation~\ref{eq:energy})
\State $\FC_{\mathrm{spec}}\gets\arg\min_{f\in\mathcal F}|\xi_a(f)-\tau|$ \Comment{No policy evaluation}
\State $\theta_{\mathrm S}\gets\theta_a$ \Comment{Copy non-target parameters from the anchor}
\For{each matrix $\ell\in\mathcal L$ of shape $m\times n$}
 \State $k\gets\max(1,\lceil\FC_{\mathrm{spec}}\min(m,n)\rceil)$; \quad $U_k\gets U^\ell_{:,1:k}$, $V_k\gets V^\ell_{:,1:k}$
 \State $\Delta_{\mathrm{prot}}\gets U_k(U_k^\top\Delta_a^\ell V_k)V_k^\top$ \Comment{Protected anchor component}
 \State $\Delta_{\mathrm{comp}}\gets\Delta_b^\ell-U_k(U_k^\top\Delta_b^\ell V_k)V_k^\top$ \Comment{Complementary donor component}
 \State $W_{\mathrm S}^\ell\gets W_0^\ell+\Delta_{\mathrm{prot}}+\Delta_{\mathrm{comp}}$
\EndFor
\State \Return $\theta_{\mathrm S}$
\end{algorithmic}
\end{algorithm}

\Needspace{5\baselineskip}
\paragraph{Computational cost.}
For $W\in\mathbb R^{m\times n}$ and $r=\min(m,n)$, a dense economy SVD costs on the order of $mnr$ operations. Evaluating $U_k(U_k^\top X V_k)V_k^\top$ avoids materializing an $(mn)\times(mn)$ projector. The resulting checkpoint has the same tensor shapes as its sources. The SVD can be reused for different FC values and for the retained-energy curve in Equation~\ref{eq:energy}. Appendix~\ref{app:compute} reports measured fusion timings and a separate reconstruction of training-step cost.

\paragraph{Boundary references.}
The entries at 0 and 1 are native checkpoint references, not an endpoint-equivalence guarantee. At zero, the code retains at least one singular direction and copies anchor non-target parameters. At one, a rectangular matrix's full economy-SVD block need not be the identity on the entire matrix space. The interior settings and delta-first projector define the analyzed method.

\paragraph{A conditional local reward criterion.}
Let $J(\theta)$ be a smooth expected-reward objective, with an $L$-Lipschitz gradient on the segment from the anchor to the fused parameters, and let $h=\theta_{\mathrm S}-\theta_a$. The standard smoothness bound gives
\begin{equation}
 J(\theta_a+h)-J(\theta_a)\geq
 \nabla J(\theta_a)^\top h-\tfrac L2\|h\|_2^2.
 \label{eq:local}
\end{equation}
If the directional benefit exceeds the curvature term, fusion improves the anchor; if the anchor is at least as strong as the donor under $J$, it exceeds both. This is a sufficient condition for a smooth objective, not a guarantee for sampled benchmark accuracy or discontinuous decoding rules. SURGE does not measure the gradient or assume that its projector enforces this condition.

\paragraph{Projection interpretation.} The fused update is the unique closest donor update whose protected component matches the anchor:
\begin{equation}
 \Delta_{\mathrm S}=\arg\min_X\|X-\Delta_b\|_F^2
 \quad\text{subject to}\quad \Keep_{a,k}(X)=\Keep_{a,k}(\Delta_a).
 \label{eq:projectionopt}
\end{equation}
\begin{proof}
The projectors satisfy $\Keep_{a,k}^2=\Keep_{a,k}$, $\Comp_{a,k}^2=\Comp_{a,k}$, and $\Keep_{a,k}\Comp_{a,k}=0$. Their ranges are perpendicular under the Frobenius inner product, so the objective decomposes into
$\|\Keep_{a,k}(X-\Delta_b)\|_F^2+\|\Comp_{a,k}(X-\Delta_b)\|_F^2$.
The constraint fixes the first term. The second has minimum zero, attained by setting
$\Comp_{a,k}(X)=\Comp_{a,k}(\Delta_b)$. Together these conditions uniquely give Equation~\ref{eq:fusion}.
\end{proof}

\clearpage
\section{Additional ablation results}
\label{app:ablations}
The ablations examine the decomposition reference, donor complement, displacement direction, protected block, and checkpoint averaging. Scalar and spectral variants use the DeepSeek sources in Section~\ref{sec:experiments}, target the same 196 body linear matrices, and keep non-target tensors from the anchor. Projector-based variants use FC $=0.8$ and the same matrix-specific $k$. Checkpoint averages instead combine all parameters over the specified windows.

\begin{table}[H]
\caption{\textbf{Expanded DeepSeek reuse controls on AIME24.} $\eta$ is normalized Frobenius distance over body linear matrices; Cap is the response-limit truncation rate. Distances are nonnegative, including for negative $\alpha$. A dash denotes an unreported distance.}
\label{tab:expanded-controls}
\centering\small
\begin{tabular*}{\linewidth}{@{\extracolsep{\fill}}lrrrr@{}}
\toprule
Variant & $\eta$ & Accuracy (\%) & Reasoning tokens & Cap (\%) \\
\midrule
Anchor-only truncation & 0.329 & 44.17 & 8,320 & 2.5 \\
\midrule
Scalar $\alpha=-0.50$ & 0.500 & 49.17 & 9,141 & 3.3 \\
Scalar $\alpha=-0.25$ & 0.250 & 47.50 & 7,723 & 0.0 \\
Scalar $\alpha=0.25$ & 0.250 & 45.00 & 8,521 & 1.7 \\
Scalar $\alpha=0.50$ & 0.500 & 44.17 & 8,028 & 0.8 \\
Scalar $\alpha=0.71$ & 0.710 & 47.50 & 7,659 & 1.7 \\
Scalar $\alpha=0.75$ & 0.750 & 46.67 & 7,275 & 0.8 \\
\midrule
Average: 3 checkpoints & --- & 50.83 & 8,078 & 0.8 \\
Average: 6 checkpoints & --- & 49.17 & 7,977 & 2.5 \\
\midrule
Trailing singular block & 0.834 & 44.17 & 7,743 & 0.8 \\
Absolute-weight SVD & 0.881 & 41.67 & 8,173 & 2.5 \\
\bottomrule
\end{tabular*}
\end{table}

\paragraph{Truncation, scalar combinations, and averaging.}
Anchor-only truncation changes the anchor by $-\Comp_{a,k}(\Delta_a)$, toward initialization in the complementary block, with normalized distance 0.329. For scalar combinations, distance is $|\alpha|$; negative coefficients extrapolate away from the donor. The best measured grid value is descriptive, not an optimum over the continuous line. The three-checkpoint average uses steps 3200/3400/3600; the six-checkpoint average uses 2600/2800/3000/3200/3400/3600. Both uniformly average all parameters. Each policy in Table~\ref{tab:expanded-controls} uses four rounds under the main AIME24 protocol.

\begin{table}[H]
\caption{\textbf{Absolute-weight versus delta-first SVD on DeepSeek.} Only the matrix used to define the singular basis changes; FC and the fusion rule are held fixed. Cells report accuracy (\%) / mean reasoning tokens.}
\label{tab:weight-delta}
\centering\small
\begin{tabular*}{\linewidth}{@{\extracolsep{\fill}}lrr@{}}
\toprule
Benchmark & Delta-SVD & Absolute-weight SVD \\
\midrule
AIME24 & \textbf{54.17} / 7,798 & 41.67 / 8,173 \\
AIME25 & \textbf{38.33} / 7,383 & 35.00 / 7,368 \\
HMMT25 & \textbf{23.33} / 7,693 & 19.17 / 7,331 \\
BRUMO25 & \textbf{51.67} / 7,285 & 50.83 / 6,941 \\
CMIMC25 & \textbf{25.63} / 8,700 & 23.75 / 8,415 \\
AMC23 & \textbf{92.50} / 5,225 & 88.75 / 4,947 \\
MATH & \textbf{92.40} / 3,415 & 91.00 / 3,199 \\
Minerva & \textbf{52.76} / 4,742 & 52.11 / 4,493 \\
Olympiad & \textbf{67.32} / 5,832 & 66.58 / 5,639 \\
\bottomrule
\end{tabular*}
\end{table}

\Needspace{5\baselineskip}
The absolute-weight variant changes only the SVD input from $W_a-W_0$ to $W_a$. The remainder of the fusion rule is identical. Delta-first SVD gives higher accuracy on all nine benchmarks, although it does not uniformly use fewer reasoning tokens than the absolute-weight variant. This comparison concerns the geometry that preserves useful RL changes; it does not assume that lower token usage by itself implies a stronger reasoning policy.

\begin{table}[H]
\caption{\textbf{Random-subspace constructions on DeepSeek AIME24.} Each repetition independently samples random left and right orthonormal bases for every target matrix at the same block size as SURGE.}
\label{tab:random-trials}
\centering\small
\begin{tabular}{@{}lrr@{}}
\toprule
Construction & Accuracy (\%) & Reasoning tokens \\
\midrule
1 & 50.83 & 7,239 \\
2 & 45.83 & 8,182 \\
3 & 47.50 & 7,205 \\
4 & 45.00 & 7,668 \\
\midrule
Mean & 47.29 & 7,574 \\
\bottomrule
\end{tabular}
\end{table}

For each repetition and target matrix of shape $m\times n$, sample Gaussian matrices of shapes $m\times k$ and $n\times k$, and take their QR factors as the left and right orthonormal bases. Each target matrix receives independently sampled bases. Four independent constructions produce the results above; they are distinct from the four sampled responses per evaluation problem. The trailing-subspace control instead uses the last $k$ singular vectors from the same delta SVD. The displacement ratio in Equation~\ref{eq:displacement} concatenates all target-matrix changes; $\alpha=0.71$ is an approximate, rounded match to the measured ratio 0.713.

\clearpage
\section{Evaluation variability}
\label{app:variability}
\begin{table}[H]
\caption{\textbf{DeepSeek variability across four evaluation rounds.} Entries give accuracy mean $\pm$ standard deviation / variance, computed across four round-level accuracy estimates. These are evaluation-sampling statistics, not variability across independent RL training runs.}
\label{tab:variance}
\centering\small
\begin{tabular*}{\linewidth}{@{\extracolsep{\fill}}lrrr@{}}
\toprule
Benchmark & SURGE & Anchor & Donor \\
\midrule
AIME24 & $\mathbf{54.17} \pm 6.40$ / 40.97 & $50.83 \pm 7.95$ / 63.19 & $47.50 \pm 5.95$ / 35.42 \\
AIME25 & $\mathbf{38.33} \pm 2.89$ / 8.33 & $36.67 \pm 4.71$ / 22.22 & $35.83 \pm 2.76$ / 7.64 \\
HMMT25 & $\mathbf{23.33} \pm 5.77$ / 33.33 & $20.00 \pm 3.33$ / 11.11 & $19.17 \pm 1.44$ / 2.08 \\
BRUMO25 & $\mathbf{51.67} \pm 1.67$ / 2.78 & $49.17 \pm 1.44$ / 2.08 & $46.67 \pm 3.33$ / 11.11 \\
CMIMC25 & $\mathbf{25.63} \pm 2.07$ / 4.30 & $22.50 \pm 4.33$ / 18.75 & $21.88 \pm 2.72$ / 7.42 \\
AMC23 & $\mathbf{92.50} \pm 3.06$ / 9.38 & $90.00 \pm 2.50$ / 6.25 & $88.13 \pm 3.70$ / 13.67 \\
MATH & $\mathbf{92.40} \pm 0.58$ / 0.34 & $91.60 \pm 0.14$ / 0.02 & $90.75 \pm 0.48$ / 0.23 \\
Minerva & $\mathbf{52.76} \pm 1.73$ / 3.01 & $52.02 \pm 0.41$ / 0.17 & $50.64 \pm 1.65$ / 2.73 \\
Olympiad & $\mathbf{67.32} \pm 0.46$ / 0.21 & $66.58 \pm 0.16$ / 0.03 & $64.80 \pm 0.74$ / 0.54 \\
\bottomrule
\end{tabular*}
\end{table}

The four round-level scores correspond to sampling one response per question in each round. Their mean is the reported four-response accuracy. Standard deviations and variances quantify variability in these round-level estimates, not variation across independent RL training runs. We report descriptive accuracy differences and do not claim per-benchmark statistical significance. The table retains rounded statistics, so squaring a displayed standard deviation need not reproduce the displayed variance exactly. These statistics describe individual cells; the pooled association across all 54 cells is analyzed below.

\subsection{Within-cell accuracy--token correlation}
\label{app:round-correlation}
The corrected run-level report contains four paired accuracy/token observations for each of 54 family--policy--benchmark cells. Let $a_{ci}$ and $t_{ci}$ be the reported accuracy and mean reasoning tokens in round $i\in\{1,\ldots,4\}$ of cell $c$. We standardize both quantities within the cell:
\begin{equation}
 z^a_{ci}=\frac{a_{ci}-\bar a_c}{s^a_c},\qquad
 z^t_{ci}=\frac{t_{ci}-\bar t_c}{s^t_c},\qquad
 r=\frac{\sum_{c=1}^{54}\sum_{i=1}^4 z^a_{ci}z^t_{ci}}{3\cdot54},
 \label{eq:pooled-correlation}
\end{equation}
where each $s_c$ is the sample standard deviation across the four rounds. All cells have nonzero variation in both metrics. Because each standardized vector has squared norm 3, Equation~\ref{eq:pooled-correlation} is both the pooled Pearson correlation and the equal-weight mean of the 54 cell-level correlations. Scaling, in addition to centering, prevents high-variance benchmarks from dominating the statistic.

For the permutation test, each replicate independently draws one of the $4!=24$ token-order permutations in every cell and recomputes $r$, leaving accuracy fixed. The null is exchangeability of the accuracy--token pairing within cells. Of 100,000 replicates, 2,062 have $|r^{\mathrm{perm}}|\geq|r^{\mathrm{obs}}|$. With the plus-one correction, the two-sided estimate is $p=(2062+1)/(100000+1)=0.02063$ (Monte Carlo standard error approximately 0.00045). The observed $r$ is $-0.18067$. The analysis script and extracted run records accompany the source. Calculations use the run accuracies as reported to one decimal place, without replacing the separately reported benchmark means.

\begin{table}[H]
\centering\small
\caption{\textbf{Within-cell accuracy--token association.} Each cell fixes family, policy, and benchmark; both metrics are standardized across its four rounds.}
\label{tab:round-correlation}
\begin{tabular}{@{}lrrr@{}}
\toprule
Family & Cells & Runs & Pooled $r$\\
\midrule
DeepSeek & 27 & 108 & $-0.142$\\
Nemotron & 27 & 108 & $-0.219$\\
\midrule
Both & 54 & 216 & $-0.181$\\
\bottomrule
\end{tabular}
\end{table}

\begin{table}[H]
\centering\small
\caption{\textbf{Within-cell correlations by benchmark.} Each family contains three policy cells (12 runs) per benchmark; the pooled column contains six cells (24 runs).}
\label{tab:correlation-benchmarks}
\begin{tabular}{@{}lrrr@{}}
\toprule
Benchmark & DeepSeek & Nemotron & Pooled\\
\midrule
AIME24 & $-0.367$ & $0.057$ & $-0.155$\\
AIME25 & $-0.494$ & $-0.194$ & $-0.344$\\
HMMT25 & $-0.086$ & $-0.151$ & $-0.119$\\
BRUMO25 & $0.190$ & $-0.524$ & $-0.167$\\
CMIMC25 & $-0.027$ & $-0.040$ & $-0.034$\\
AMC23 & $-0.208$ & $-0.791$ & $-0.500$\\
MATH & $-0.266$ & $0.002$ & $-0.132$\\
Minerva & $-0.032$ & $-0.492$ & $-0.262$\\
Olympiad & $0.010$ & $0.161$ & $0.085$\\
\bottomrule
\end{tabular}
\end{table}

Eight of nine pooled benchmark correlations are negative; Olympiad is the exception. The family-specific columns also show local sign variation, so the pooled trend is not a universal within-benchmark rule. These are aggregate round-level associations under fixed evaluation settings. They do not show that a correct response is shorter than an incorrect response to the same question, or that imposing a shorter generation causes higher accuracy.

\clearpage
\section{Controller measurements}
\label{app:sweeps}
\begin{table}[H]
\caption{\textbf{Full controller sweeps on AIME24.} FC $=0$ and $1$ denote native donor and anchor reference positions. Tokens are mean reasoning tokens.}
\label{tab:sweep}
\centering\small
\begin{tabular*}{\linewidth}{@{\extracolsep{\fill}}lrrrr@{}}
\toprule
& \multicolumn{2}{c}{DeepSeek} & \multicolumn{2}{c}{Nemotron} \\
FC / reference & Accuracy (\%) & Tokens & Accuracy (\%) & Tokens \\
\midrule
0.0 (donor) & 47.50 & 7,651 & 60.00 & 9,212 \\
0.1 & 47.50 & 7,517 & 62.50 & 10,060 \\
0.2 & 48.33 & 7,639 & 65.00 & 10,420 \\
0.3 & 49.17 & 7,439 & 66.67 & 10,786 \\
0.4 & 48.33 & 8,096 & 66.67 & 10,841 \\
0.5 & 49.17 & 7,795 & 67.50 & 11,159 \\
0.6 & 50.00 & 7,864 & 65.83 & 11,236 \\
0.7 & 51.67 & 7,633 & 68.33 & 11,285 \\
0.8 & \textbf{54.17} & 7,798 & \textbf{70.00} & 11,395 \\
0.9 & 50.83 & 7,773 & 68.33 & 11,198 \\
1.0 (anchor) & 50.83 & 8,065 & 68.33 & 13,030 \\
\bottomrule
\end{tabular*}
\end{table}

\begin{figure}[H]
\centering
\includegraphics[width=\linewidth]{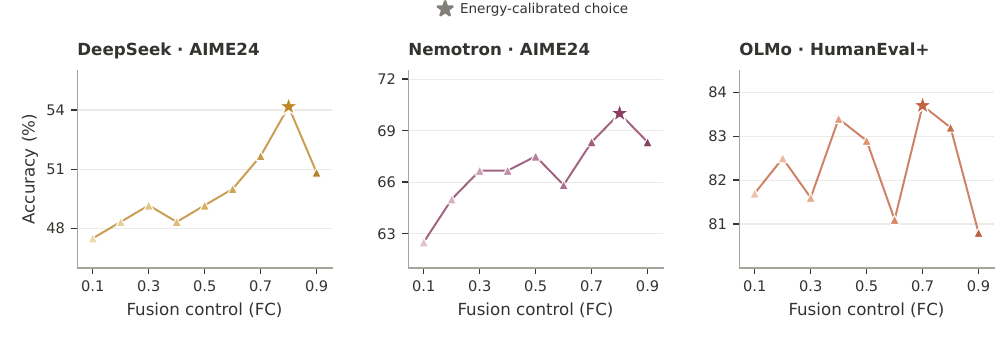}
\caption{\textbf{Accuracy in raw FC coordinates.} Stars denote the energy-calibrated settings. All nine interior settings per family use four evaluation rounds. Native reference positions are omitted from the curves.}
\label{fig:raw-fc}
\end{figure}
\begin{table}[H]
\centering\small
\caption{\textbf{Complete HumanEval+ sensitivity measurements.} Every setting uses four evaluation rounds; $\eta$ is measured fusion displacement. MBPP+ is confined to the main-policy comparison.}
\label{tab:coding-sweep}
\begin{tabular}{@{}lrrrr@{}}
\toprule
FC & Retained energy $\xi_a$ & Displacement $\eta$ & Accuracy (\%) & Tokens\\
\midrule
0.1 & 0.509 & 0.929 & 81.7 & 4,258 \\
0.2 & 0.655 & 0.886 & 82.5 & 4,146 \\
0.3 & 0.754 & 0.846 & 81.6 & 4,338 \\
0.4 & 0.828 & 0.807 & 83.4 & 4,462 \\
0.5 & 0.883 & 0.766 & 82.9 & 4,498 \\
0.6 & 0.924 & 0.724 & 81.1 & 4,591 \\
\textbf{0.7} & 0.954 & 0.680 & \textbf{83.7} & 4,732 \\
0.8 & 0.976 & 0.632 & 83.2 & 4,791 \\
0.9 & 0.991 & 0.582 & 80.8 & 4,939 \\
\bottomrule
\end{tabular}
\end{table}

\clearpage
\section{Qualitative reasoning examples}
\label{app:cases}
The four exact-answer cases below compare representative math responses with the source and fused policies side by side. A coding example supplements the HumanEval+ comparison in Table~\ref{tab:case-code-chars}. Each correctness count is over four responses per policy. Reasoning-token counts belong to the displayed individual response, not the four-response mean. Cases are selected to make specific success and failure patterns visible, rather than estimate their prevalence. The rounding example isolates the final classification step that distinguishes the selected responses.

\begin{table}[H]
\caption{\textbf{DeepSeek / MATH \#491: counting.} Representative reasoning summaries and four-response correctness counts.}
\label{tab:case-counting}
\small
\begin{tabularx}{\linewidth}{@{}Y Y@{}}
\toprule
\multicolumn{2}{@{}p{\linewidth}@{}}{\textbf{Problem.} Mr.\ Potato Head has three hairstyles (or can be bald), two eyebrow pairs, one eye pair, two ear pairs, two lip pairs, and two alternative shoe pairs. All facial parts and one complete shoe pair are required. How many appearances are possible? \quad\textbf{Ground truth:} $64$.}\\
\midrule
\textbf{Donor} & \textbf{Anchor}\\
\midrule
\textcolor{surgered}{Treats mandatory paired parts as optional left/right choices and uses an unsupported product of options. The resulting count does not respect the specified complete pairs.} & \textcolor{surgered}{Recognizes baldness but inconsistently counts the remaining features and shoe alternatives. Its final product undercounts the valid combinations.}\\[4pt]
Answer: $162$ \textcolor{surgered}{(incorrect)} & Answer: $32$ \textcolor{surgered}{(incorrect)}\\
Selected response: 2,677 reasoning tokens & Selected response: 3,013 reasoning tokens\\
Observed correct: 0/4 & Observed correct: 0/4\\
\midrule
\multicolumn{2}{@{}l@{}}{\textbf{SURGE, FC = 0.8}}\\
\multicolumn{2}{@{}p{\linewidth}@{}}{\textcolor{surgeteal}{Includes baldness as a fourth hair choice and treats the two complete shoe pairs as alternatives. It then combines the independent choices without adding optional missing facial parts.}}\\[4pt]
\multicolumn{2}{@{}c@{}}{$\displaystyle 4\cdot2\cdot1\cdot2\cdot2\cdot2=64.$}\\[4pt]
Answer: $64$ \textcolor{surgeteal}{(correct)} & Observed correct: \textbf{3/4}\\
\multicolumn{2}{@{}l@{}}{Selected response: 4,098 reasoning tokens}\\
\bottomrule
\end{tabularx}
\end{table}

\begin{table}[H]
\caption{\textbf{Nemotron / MATH \#119: parentheses.} Representative reasoning summaries and four-response correctness counts.}
\label{tab:case-parentheses}
\small
\begin{tabularx}{\linewidth}{@{}Y Y@{}}
\toprule
\multicolumn{2}{@{}p{\linewidth}@{}}{\textbf{Problem.} Insert parentheses in $1+2+3-4+5+6$ to minimize its value, preserving every number, operator, and their order. \quad\textbf{Ground truth:} $-9$.}\\
\midrule
\textbf{Donor} & \textbf{Anchor}\\
\midrule
\textcolor{surgered}{Introduces a subtraction inside a group containing 5 and 6, changing the original plus sign. Its candidate therefore solves a different expression.} & \textcolor{surgered}{Stops at the example value 3 after considering restricted groupings, while claiming that all possibilities have been checked. It fails to place the entire suffix under the minus sign.}\\[4pt]
Answer: $1$ \textcolor{surgered}{(incorrect)} & Answer: $3$ \textcolor{surgered}{(incorrect)}\\
Selected response: 6,988 reasoning tokens & Selected response: 10,339 reasoning tokens\\
Observed correct: 1/4 & Observed correct: 1/4\\
\midrule
\multicolumn{2}{@{}l@{}}{\textbf{SURGE, FC = 0.8}}\\
\multicolumn{2}{@{}p{\linewidth}@{}}{\textcolor{surgeteal}{Groups all terms after the original minus sign together. This preserves the expression and subtracts the largest possible positive suffix, producing the minimum.}}\\[4pt]
\multicolumn{2}{@{}c@{}}{$\displaystyle (1+2+3)-(4+5+6)=6-15=-9.$}\\[4pt]
Answer: $-9$ \textcolor{surgeteal}{(correct)} & Observed correct: \textbf{3/4}\\
\multicolumn{2}{@{}l@{}}{Selected response: 7,865 reasoning tokens}\\
\bottomrule
\end{tabularx}
\end{table}

\clearpage
\begin{table}[H]
\caption{\textbf{Nemotron / Olympiad \#544: incenters.} Representative reasoning summaries and four-response correctness counts.}
\label{tab:case-incenters}
\small
\begin{tabularx}{\linewidth}{@{}Y Y@{}}
\toprule
\multicolumn{2}{@{}p{\linewidth}@{}}{\textbf{Problem.} A square $ABCD$ has diagonal 1. Points $E\in AB$ and $F\in BC$ satisfy $\angle BCE=\angle BAF=30^\circ$, and $G=CE\cap AF$. Find the distance between the incenters of $AGE$ and $CGF$. \quad\textbf{Ground truth:} $4-2\sqrt3$.}\\
\midrule
\textbf{Donor} & \textbf{Anchor}\\
\midrule
\textcolor{surgered}{Misreads the angle condition as involving the square diagonal, declares a conflict with its $45^\circ$ angle, and then appeals to unsupported symmetry to identify the two incenters.} & \textcolor{surgered}{Uses coordinates and an incenter calculation, but errors in the subsequent coordinate arithmetic lead to an incorrect distance.}\\[4pt]
Answer: $0$ \textcolor{surgered}{(incorrect)} & Answer: $1/2$ \textcolor{surgered}{(incorrect)}\\
Selected response: 10,119 reasoning tokens & Selected response: 17,122 reasoning tokens\\
Observed correct: 0/4 & Observed correct: 0/4\\
\midrule
\multicolumn{2}{@{}l@{}}{\textbf{SURGE, FC = 0.8}}\\
\multicolumn{2}{@{}p{\linewidth}@{}}{\textcolor{surgeteal}{Carries the coordinate method through both incenter calculations. With $A=(0,0)$ and $B=(1/\sqrt2,0)$, the two incenter coordinates differ by $2\sqrt2-\sqrt6$ in each axis, giving the required distance.}}\\[4pt]
\multicolumn{2}{@{}c@{}}{$\displaystyle \|I_{AGE}-I_{CGF}\|=\sqrt{2}\,(2\sqrt2-\sqrt6)=4-2\sqrt3.$}\\[4pt]
Answer: $4-2\sqrt3$ \textcolor{surgeteal}{(correct)} & Observed correct: \textbf{2/4}\\
\multicolumn{2}{@{}l@{}}{Selected response: 19,582 reasoning tokens}\\
\bottomrule
\end{tabularx}
\end{table}

\begin{table}[H]
\caption{\textbf{Nemotron / Olympiad \#480: rounding.} Representative reasoning summaries and four-response correctness counts.}
\label{tab:case-rounding}
\small
\begin{tabularx}{\linewidth}{@{}Y Y@{}}
\toprule
\multicolumn{2}{@{}p{\linewidth}@{}}{\textbf{Problem.} Find the largest real solution of $\lfloor x/3\rfloor+\lceil3x\rceil=\sqrt{11}\,x$. \quad\textbf{Ground truth:} $189\sqrt{11}/11$.}\\
\midrule
\textbf{Donor} & \textbf{Anchor}\\
\midrule
\textcolor{surgered}{Tests isolated candidates and extrapolates an unsupported arithmetic pattern. Failure of a later candidate is treated as evidence that the preceding one is maximal.} & \textcolor{surgered}{Sets up the piecewise parametrization, but decimal-division errors underestimate the admissible integer ranges. It stops at a valid, nonmaximal solution.}\\[4pt]
Answer: $29/\sqrt{11}$ \textcolor{surgered}{(incorrect)} & Answer: $84/\sqrt{11}$ \textcolor{surgered}{(incorrect)}\\
Selected response: 14,224 reasoning tokens & Selected response: 17,295 reasoning tokens\\
Observed correct: 0/4 & Observed correct: 0/4\\
\midrule
\multicolumn{2}{@{}l@{}}{\textbf{SURGE, FC = 0.8}}\\
\multicolumn{2}{@{}p{\linewidth}@{}}{\textcolor{surgeteal}{Writes $x=3m+r$ with $0\leq r<3$ and considers $k=\lceil3r\rceil$. The final case $k=9$ permits $m=18$. Completing this case gives the largest solution; direct substitution yields floor 18 and ceiling 171.}}\\[4pt]
\multicolumn{2}{@{}c@{}}{$\displaystyle x=\frac{10m+k}{\sqrt{11}}=\frac{189}{\sqrt{11}}=\frac{189\sqrt{11}}{11}.$}\\[4pt]
Answer: $189\sqrt{11}/11$ \textcolor{surgeteal}{(correct)} & Observed correct: \textbf{2/4}\\
\multicolumn{2}{@{}l@{}}{Selected response: 12,238 reasoning tokens}\\
\bottomrule
\end{tabularx}
\end{table}

\paragraph{Why the larger answer is valid.}
With $x=3m+r$ and $k=\lceil3r\rceil$, the equation becomes $10m+k=\sqrt{11}\,x$. Feasibility gives $m\leq2k$ for $1\leq k\leq8$ and $m\leq18$ for $k=9$; the latter yields $x=189/\sqrt{11}$. Its floor and ceiling terms are 18 and 171. Completing this final case, rather than stopping at a smaller feasible candidate, is the decisive mathematical difference.

\clearpage
\subsection{Coding examples: specification fidelity}
\label{app:code-cases}
The HumanEval+ example in Table~\ref{tab:case-code-chars} and the MBPP+ example below use the energy-calibrated OLMo policy (FC $=0.7$) and its sources, with four responses per problem. Correctness means passing both the original and additional EvalPlus tests. The tables summarize submitted programs rather than full reasoning traces. They were selected to expose concrete failure modes; they do not estimate how often fusion fixes or introduces errors across the benchmark.

\begin{table}[H]
\caption{\textbf{OLMo / MBPP+ \#792: counting lists.} SURGE distinguishes the requested type predicate from the shortcut suggested by homogeneous examples.}
\label{tab:case-code-lists}
\small
\begin{tabularx}{\linewidth}{@{}Y Y@{}}
\toprule
\multicolumn{2}{@{}p{\linewidth}@{}}{\textbf{Task.} Implement \texttt{count\_list(lst)} to count top-level elements that are lists. For mixed inputs, counting all elements is incorrect.}\\
\midrule
\textbf{Donor} & \textbf{Anchor}\\
\midrule
\textcolor{surgered}{Three responses return the outer length. One uses the correct type test but submits the wrong function name, \texttt{count\_lists}, and fails the required interface.} & \textcolor{surgered}{Three responses return the outer length, fitting the all-list examples but failing mixed-type inputs. Only one tests whether each element is a list.}\\[4pt]
Observed correct: 0/4 & Observed correct: 1/4\\
\midrule
\multicolumn{2}{@{}l@{}}{\textbf{SURGE}}\\
\multicolumn{2}{@{}p{\linewidth}@{}}{\textcolor{surgeteal}{Three responses use \texttt{isinstance(item, list)} under the required function name. They count list-valued elements while ignoring scalars. One response still uses the outer-length shortcut.}}\\[4pt]
\multicolumn{2}{@{}c@{}}{\texttt{return sum(1 for item in lst if isinstance(item, list))}}\\[4pt]
\multicolumn{2}{@{}l@{}}{Observed correct: \textbf{3/4}}\\
\bottomrule
\end{tabularx}
\end{table}

\clearpage
\section{Spectral calibration of RL updates}
\label{app:spectral}
The diagnostics below use the same anchor deltas and target matrices as the main fusions: 196 matrices each for DeepSeek and Nemotron, and 224 for OLMo. Nemotron's anchor is step 3440; the energy curve here is measured on that checkpoint. All quantities are computed from weights without generated responses. Equation~\ref{eq:energy} aggregates squared singular values before taking the ratio, weighting matrices by their update energy.

\subsection{One target, different rank fractions}
\begin{table}[H]
\centering\small
\caption{\textbf{Retained anchor-update energy $\xi_a$ across FC.} Bold entries are nearest to $\tau=0.95$. The Nemotron curve uses step 3440.}
\label{tab:spectral-energy}
\setlength{\tabcolsep}{4pt}
\begin{tabular*}{\linewidth}{@{\extracolsep{\fill}}lrrrrrrrrr@{}}
\toprule
FC & 0.1 & 0.2 & 0.3 & 0.4 & 0.5 & 0.6 & 0.7 & 0.8 & 0.9\\
\midrule
DeepSeek & 0.518 & 0.638 & 0.724 & 0.793 & 0.848 & 0.893 & 0.929 & \textbf{0.959} & 0.983 \\
Nemotron & 0.405 & 0.546 & 0.652 & 0.736 & 0.805 & 0.862 & 0.909 & \textbf{0.947} & 0.978 \\
OLMo & 0.509 & 0.655 & 0.754 & 0.828 & 0.883 & 0.924 & \textbf{0.954} & 0.976 & 0.991 \\
\bottomrule
\end{tabular*}
\end{table}

Applying Equation~\ref{eq:fc-selection} gives FC $=0.8/0.8/0.7$ for DeepSeek/Nemotron/OLMo. Their distances from $\tau=0.95$ are $0.009/0.003/0.004$. The rule picks the nearest candidate, not the first candidate above the target; Nemotron's selected retention is slightly below 0.95. Values are rounded for reporting. No interpolation of the accuracy measurements is used to select FC.

The target was identified from the initial math study and fixed before applying the procedure to OLMo. All three trajectories then use the same construction: compute the anchor RL delta, obtain its retained-energy curve, select the nearest candidate to the shared target, and fuse once. Conditional on the source pair and fixed target, this procedure uses weights alone and requires no accuracy sweep. Its agreement with strong policies across math and code does not establish a universal optimal energy level. The OLMo finer-grid diagnostics, $\xi_a(0.65)=0.940$ and $\xi_a(0.75)=0.966$, bracket the selected value; no behavioral measurements are available at those two settings. The four-round sweeps characterize sensitivity (Appendix~\ref{app:sweeps}).

\subsection{Why equal rank fractions preserve different energy}
DeepSeek and Nemotron have 28 layers, with $q/o$ matrices of shape $1536\times1536$, $k/v$ of shape $256\times1536$, and MLP matrices of shape $1536\times8960$ or its transpose. OLMo has 32 layers, square $4096\times4096$ attention projections, and $4096\times11008$ MLP projections or their transposes. The same retained rank fraction therefore need not capture the same energy even in a shape-matched Gaussian reference.

\begin{table}[H]
\centering\small
\caption{\textbf{Actual versus shape-matched reference retention.} Each reference is weighted by the corresponding anchor's module energies.}
\label{tab:spectral-shape}
\begin{tabular}{@{}lrrrr@{}}
\toprule
 & \multicolumn{2}{c}{FC $=0.7$} & \multicolumn{2}{c}{FC $=0.8$}\\
\cmidrule(lr){2-3}\cmidrule(l){4-5}
Anchor & Actual & Reference & Actual & Reference\\
\midrule
DeepSeek & .929 & .854 & .959 & .913\\
Nemotron & .909 & .854 & .947 & .913\\
OLMo & .954 & .920 & .976 & .958\\
\bottomrule
\end{tabular}
\end{table}

\Needspace{6\baselineskip}
The reference in Table~\ref{tab:spectral-shape} uses same-shape Gaussian matrices and weights their retention by the corresponding anchor's module energy. At FC $=0.8$, the OLMo reference already retains 95.8\%, compared with 91.3\% for the math architectures. Its actual update retains 97.6\%. Matching $\xi_a$ adjusts for the observed energy distribution rather than assuming that raw FC denotes the same protected content across models. It does not remove every architectural or donor-dependent difference.

\subsection{Protected energy is distinct from fusion displacement}
Retained energy depends only on the anchor. In contrast, the normalized displacement $\eta$ in Equation~\ref{eq:displacement} depends on the donor as well. At the selected settings, $\eta$ is 0.713, 0.559, and 0.680 for DeepSeek, Nemotron, and OLMo. Thus, retaining about 95\% of anchor energy does not mean moving 5\% of the way to the donor, nor does it predict the donor's behavioral contribution.

The unprotected anchor tails contain 4.1\%, 5.3\%, and 4.6\% of update energy, respectively; 98\%, 98\%, and 95\% of these tails come from MLP projections. With the diagnostic spike threshold set at 1.05 times the same-shape Gaussian spectral edge (scaled to match the median singular value), every selected cutoff falls after the spikes (0/196, 0/196, and 0/224 cutoffs fall within them). These are descriptive spectral diagnostics. A low-energy tail is not necessarily behaviorally irrelevant, and the 95\% target is not an intrinsic spectral boundary. In particular, OLMo's residual $q/k$ spectra need not resemble the Gaussian reference.

\subsection{Absolute-weight blocks versus update-defined blocks}
Let $\Keep^{\mathrm{abs},\ell}_{a,k}$ use the leading singular vectors of the absolute anchor weights $W_a^\ell$. Its captured RL-update energy is
\begin{equation}
 \zeta_{\mathrm{abs}}(\FC)=
 \frac{\sum_{\ell\in\mathcal L}\|\Keep^{\mathrm{abs},\ell}_{a,k_\ell}(\Delta_a^\ell)\|_F^2}
 {\sum_{\ell\in\mathcal L}\|\Delta_a^\ell\|_F^2}.
 \label{eq:absolute-energy}
\end{equation}
For delta-defined blocks this equals $\xi_a(\FC)$. Table~\ref{tab:spectral-overlap} isolates the comparison for the DeepSeek pair used in the accuracy ablation. Equal block dimensions preserve very different amounts of the RL update, connecting the weight-space diagnostic to the controlled behavioral comparison.

\begin{table}[H]
\centering\small
\caption{\textbf{DeepSeek RL-update energy captured at equal block size (\%).} The source pair is the one used for the main accuracy ablation.}
\label{tab:spectral-overlap}
\begin{tabular}{@{}lrr@{}}
\toprule
FC & Absolute-weight SVD & Delta-SVD\\
\midrule
0.2 & 7.0 & 63.8\\
0.8 & 28.5 & 95.9\\
\bottomrule
\end{tabular}
\end{table}

\clearpage
\section{Offline fusion cost and training-step comparison}
\label{app:compute}
\subsection{Measured fusion time}
The benchmark uses one RTX A6000 with 48 GB memory, PyTorch 2.6, and FC $=0.8$. The timed pairs are DeepSeek steps 3600/2600 and Nemotron steps 3440/1200, matching the main math experiments. No 7B latency is inferred from these measurements. A complete pass reads the three checkpoints, computes FP32 delta SVDs and projections for 196 target matrices, evaluates the $\xi$ and $\eta$ diagnostics, and writes BF16 safetensors. Table~\ref{tab:fusion-runtime} separates measured warm-cache times from storage and CPU estimates.

\begin{table}[H]
\centering\small
\caption{\textbf{One fusion pass on one RTX A6000 (seconds).} Warm-cache entries are measured. The final two rows are estimates, not end-to-end measurements.}
\label{tab:fusion-runtime}
\begin{tabular*}{\linewidth}{@{\extracolsep{\fill}}lrr@{}}
\toprule
Stage & DeepSeek & Nemotron \\
\midrule
Read three checkpoints (warm page cache) & 0.04 & 0.03 \\
Transfer to GPU & 1.38 & 1.40 \\
196 FP32 SVDs & 29.45 & 29.56 \\
Projection and fusion & 1.81 & 1.79 \\
Retained-energy curve $\xi$ & 0.07 & 0.07 \\
Displacement curve $\eta$ & 0.34 & 0.34 \\
Write BF16 safetensors & 4.10 & 6.95 \\
\midrule
\textbf{End-to-end, warm cache} & \textbf{37.2} & \textbf{40.1} \\
\midrule
Cold-storage estimate (240 MB/s) & 83 & 80 \\
CPU-SVD estimate (48 threads; one-layer extrapolation) & 128 & 124 \\
\bottomrule
\end{tabular*}
\end{table}

The $\xi$ curve reuses the singular values and takes 0.07 seconds for either family; evaluating the reported $\eta$ curve adds 0.34 seconds. Spectral calibration therefore does not require repeated SVDs or an inference sweep. Given an FC value, only one fused checkpoint is needed. A different FC can reuse the decompositions, although constructing and evaluating additional policies still incurs projection, writing, and inference costs. These timings are individual benchmark measurements, not variability estimates.

\subsection{Reconstructing one RL step}
We separately measure rollout and actor throughput on the same GPU and reconstruct the resource cost of the JustRL recipe. This is not an end-to-end training run. One step contains $256\times8=2048$ rollouts, with temperature 1.0 and at most 15,360 response tokens. GRPO uses no critic or KL-reference pass. Actor measurements use FP32 master weights, BF16 autocast, gradient checkpointing, FlashAttention 2.7.4, and micro-batch size one; four mini-batches of 64 prompts yield four optimizer updates. Rollouts use vLLM 0.8.4 and DAPO-Math-17k prompts.

\begin{table}[H]
\centering\small
\caption{\textbf{Reconstructed RL-step resource cost (A6000-hours).} Components use single-GPU measurements; distributed overheads are excluded. Totals are as reported; displayed components are rounded.}
\label{tab:rl-step-cost}
\begin{tabular*}{\linewidth}{@{\extracolsep{\fill}}lrr@{}}
\toprule
Component / reconstruction & DeepSeek & Nemotron \\
\midrule
Rollout, 32-way synchronous accounting & 1.60 & 2.46 \\
Rollout, larger-batch throughput accounting & 0.76 & 1.76 \\
Old-policy log probabilities & 0.14 & 0.32 \\
Actor forward/backward and four AdamW updates & 0.57 & 1.24 \\
\midrule
Total, synchronous & 2.32 & 4.01 \\
Total, throughput-based & 1.48 & 3.31 \\
\bottomrule
\end{tabular*}
\end{table}

The synchronous reconstruction multiplies the single-GPU wall time for 64 responses by 32, reflecting the 32-way allocation of a 2048-response step. Those measured times are 180/277 seconds for DeepSeek/Nemotron. A separate 256-response measurement gives 3,327/3,094 generated tokens per second and mean response lengths 4,472/9,579; mean prompt lengths are 166/174. The throughput-based estimate uses these larger-batch rates instead of the 64-response wall time. Forward and backward token throughput and four AdamW updates supply the actor terms.

Both reconstructions omit weight synchronization, distributed communication, reward scoring, data loading, periodic validation, and training-checkpoint I/O. They are lower-bound accounting estimates under the stated execution assumptions, not measured distributed-training costs. In particular, the synchronous estimate is not a hardware-independent lower bound for asynchronous rollout systems.

\paragraph{Interpreting the ratio.}
The warm-cache fusion resource costs are $37.2/3600=0.0103$ and $40.1/3600=0.0111$ A6000-hours. They are 0.45\%/0.28\% of the synchronous reconstruction and 0.70\%/0.34\% of the throughput-based reconstruction. These compare aggregate GPU resource use, not the wall time of one GPU against 32 GPUs.

\paragraph{Evaluation is a separate cost.}
The timings above cover fusion and weight-space diagnostics, not accuracy evaluation. The included spectral curve supports the weight-only FC rule; evaluating candidate policies is a separate cost. The complete sweeps in this paper characterize sensitivity; they are not mandatory for every checkpoint pair. Deployment can start from a geometrically calibrated candidate and evaluate alternatives as needed on development data. Accuracy evaluation depends on candidate count and decoding protocol; these timings do not measure end-to-end search under the paper's four-response protocol.

\clearpage
\section{7B coding setup and checkpoint measurements}
\label{app:coding}
\paragraph{Training and initialization.}
The RL initialization is Olmo-3-1025-7B. Coding RL follows the OlmoRL recipe on Dolci-RL-Zero-Code-7B, using execution-based verification and a GRPO-derived objective \citep{olmo2025olmo3}. The recipe uses 32 prompts per batch, eight responses per prompt, temperature 1.0, a constant learning rate of $10^{-6}$, and maximum prompt/response lengths of 2K/16K. It uses active sampling after filtering groups with identical rewards, token-level loss normalization, and asymmetric clipping with lower/upper widths 0.2/0.272. The 3.1 recipe retains truncated sequences in the training loss.

\paragraph{Evaluation.}
HumanEval+ contains 164 problems and MBPP+ contains 378. Decoding uses temperature 1.0, top-$p$ 1.0, and a 32K response limit. Each main policy is evaluated in four rounds with one response per problem per round. The reported accuracy is the mean single-response pass rate, requiring both base and additional EvalPlus tests to pass; it is not success among four attempts. Tokens average the full generated responses. These settings differ from the math decoding configuration, but are fixed across coding sources and fusions.

\paragraph{Checkpoint roles and coverage.}
Table~\ref{tab:code-trajectory} covers 19 measured checkpoints from initialization (step 0) through step 1950. HumanEval+ is evaluated throughout; both coding benchmarks are evaluated for the six later candidates at steps 1300--1700 and 1950. Among these six, step 1950 has the highest two-benchmark mean accuracy and serves as the anchor. Step 1500 supplies a competitive donor: it has the highest native HumanEval+ accuracy and shorter responses on both tasks. These source-policy measurements determine the roles; the fixed spectral target determines FC. The trajectory and sensitivity analyses use HumanEval+ only.

\begin{table}[H]
\centering\small
\caption{\textbf{Native coding trajectory on HumanEval+.} Every checkpoint uses four evaluation rounds. SD is the reported standard deviation across round accuracies.}
\label{tab:code-trajectory}
\begin{tabular}{@{}llrrr@{}}
\toprule
RL step & Role & Accuracy (\%) & SD & Reasoning tokens\\
\midrule
0 & Initialization & 42.1 & 3.3 & 2,293 \\
100 &  & 65.1 & 2.3 & 5,090 \\
200 &  & 67.2 & 4.1 & 4,521 \\
300 &  & 70.6 & 3.7 & 4,597 \\
400 &  & 75.0 & 3.6 & 4,609 \\
500 &  & 74.7 & 1.5 & 4,317 \\
600 &  & 76.1 & 2.1 & 4,025 \\
700 &  & 76.2 & 3.1 & 4,320 \\
800 &  & 77.9 & 3.2 & 4,220 \\
900 &  & 79.3 & 1.7 & 4,457 \\
1000 &  & 79.1 & 1.8 & 4,430 \\
1100 &  & 80.0 & 1.5 & 3,892 \\
1200 &  & 80.2 & 2.1 & 4,252 \\
1300 &  & 79.3 & 2.5 & 3,877 \\
1400 &  & 79.9 & 1.0 & 4,372 \\
1500 & Donor & 82.8 & 1.4 & 4,173 \\
1600 &  & 79.7 & 2.4 & 3,932 \\
1700 &  & 81.6 & 1.0 & 4,271 \\
1950 & Anchor & 82.0 & 0.4 & 5,275 \\
\bottomrule
\end{tabular}
\end{table}

\paragraph{Reported precision.}
Coding accuracies retain the one-decimal precision of the result report. Changes in Table~\ref{tab:coding} are computed from those displayed means. The 0.0-point MBPP+ change therefore denotes equality at reported precision. The four HumanEval+ round accuracies are 84.1/83.5/83.5/83.5 for SURGE, 82.3/81.7/82.3/81.7 for the anchor, and 84.1/81.1/82.3/83.5 for the donor. All nine FC settings use four rounds (Table~\ref{tab:coding-sweep}); no intermediate scores are imputed. Standard deviations retain the source report's rounding.

\end{document}